# Drone Path-Following in GPS-Denied Environments using Convolutional Networks

M. Samy, K. Amer, M. Shaker and M. ElHelw
Center for Informatics Science
Nile University
Giza, Egypt
{k.amer, m.serag, melhelw}@nu.edu.eg

## Abstract

*This paper presents a simple approach for drone navigation to follow a predetermined path using visual input only without reliance on a Global Positioning System (GPS). A Convolutional Neural Network (CNN) is used to output the steering command of the drone in an end-to-end approach. We tested our approach in two simulated environments in the Unreal Engine using the AirSim plugin for drone simulation. Results show that the proposed approach, despite its simplicity, has average cross track distance less than 2.9 meters in the simulated environment. We also investigate the significance of data augmentation in path following. Finally, we conclude by suggesting possible enhancements for extending our approach to more difficult paths in real life, in the hope that one day visual navigation will become the norm in GPS-denied zones.*

## 1. Introduction

Quadrotors are used in many applications due to their small size and being cheaper over the past years. They are used for delivery, monitoring, and photography. They are equipped with sensors like cameras, Inertial Measurement Unit (IMU), and Global Positioning System (GPS) to increase their autonomy. One of the main tasks that drones are demanded to perform robustly is path following where a path is specified in terms of waypoints and the drone is required to cover all those waypoints. The waypoints are usually entered by selecting geolocations from an offline map of the area above which the drone will be flying. Figure 1 shows a top-down view of the Landscape Mountains environment in Unreal Engine in the top image, the path is drawn in blue points, the bottom image shows a success case where the path is covered by the drone during testing.

Path following is typically implemented using Global Positioning System (GPS) for localization integrated with closed-loop control using IMU. The feedback from the IMU and GPS allow the drone to move along its path with minimal error. However, when GPS signal is weak or not available the drone suffers from the drift problem due to the accumulative error. Reliance on GPS in this approach has many weaknesses: as mentioned GPS is a remote signal, it is not guaranteed to be received as in GPS-denied environments like indoors environments or in areas surrounded by high buildings, even when the drone is in line-of-sight with the satellite sending the signal, the information in the remote signal could be modified by some attackers which is known as GPS spoofing. Finally, being a radio signal, it is susceptible to signal interference, in practice the last issue is not a major weakness relative to GPS denied environments and GPS spoofing.

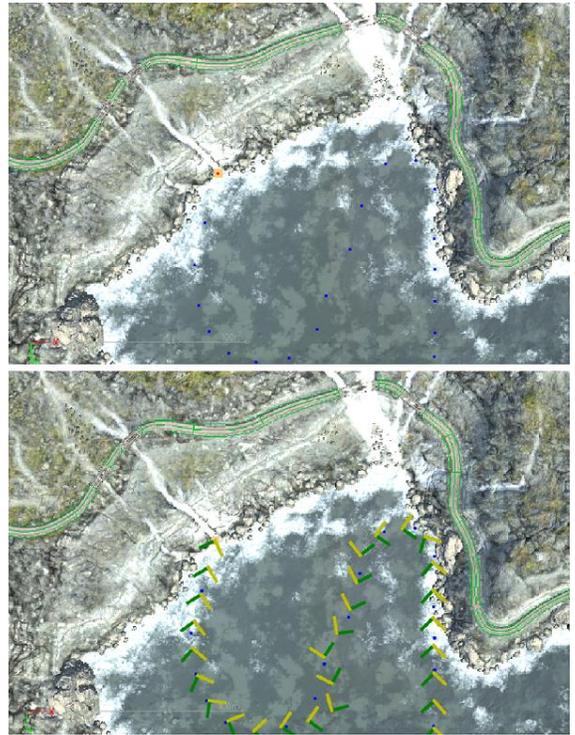

Figure 1. Path determined in blue dots, red dot shows the starting point (top), path that is followed by the drone during testing (bottom).

We propose using visual input for path following instead of relying on GPS for localization and IMU for control. Visual input is collected by an onboard camera on the drone, thus it mitigates all the mentioned weaknesses of GPS. Visual information was used for air navigation before the invention of GPS as there exists discriminative visual features that can be correlated with the waypoints in a path, thus a pilot can localize himself with respect to known landmarks along his path and take actions accordingly. This approach requires an offline map to extract the discriminative landmarks and pre-flight training to correlate the visual landmark with the correct action to take in order to follow the specified path. In our approach we use a CNN to extract semantic features automatically from the visual input followed by a two layer fully connected neural network to correlate those features with the correct action.

To test our approach, we used Unreal Engine 4[1], a game engine that is used to build games and simulations. We used two environments: Blocks and Landscape Mountains, an environment is a 3D model that we use in place of the offline map. Figure 2 shows some sample images from both environments, Blocks environment contains same sized cubes that are placed above each other and each group has a different color, Landscape Mountains environment is a more complex environment containing more natural scenes of frozen lakes, trees, and mountains. We started by Blocks as it is faster in terms of frames per second and has more discriminative features based on color and shapes of each blocks group. Then we moved to Landscape Mountains to show that our approach is working in more realistic scenes. We used AirSim [1], an Unreal plugin for the simulation of the quadrotor, it provides an API for the control of the quadrotor inside the environments besides collecting visual data for training.

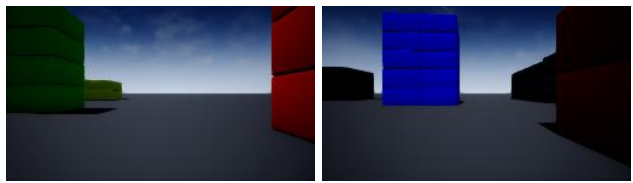

---

[1] https://www.unrealengine.com

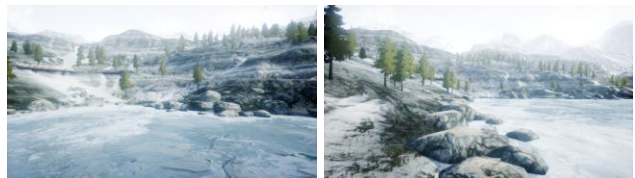

Figure 2. Sample images taken by the drone's camera. Blocks (top) Mountains Landscape (bottom).

## 2. Related Work

CNNs have shown outstanding results in the problem of classification at ImageNet Large Scale Visual Recognition Competition (ILSVRC) [2], they are the state-of-the-art models since 2012 for this challenge. From there on, CNNs have been used in other vision tasks such as localization [3], detection [4], [5], and segmentation [6], [7]. Their performance in these tasks also outperformed traditional computer vision algorithms. Only recently, they have been used in an end-to-end manner in navigation using visual input, as in [8] where a deep CNN is used to map the input image from a front facing camera to a steering angle command, the network uses the visual input to associate straight road with a small angle and turning segments of the road with a convenient angle to keep the car in the lane.

Attaining autonomous behavior for UAVs and robots is an active research area. One direction for solving such a problem is using Reinforcement Learning (RL) to learn a policy for the control. Zhang et. al. [9] combined RL with model predictive control to train a deep neural network on obstacle avoidance when deployed on a UAV. Zho et. al. [10] proposed a RL based model that navigates in an indoor scene in order to search for a target object. Their Siamese model takes an input the image of target object and a scene observation. Chaplot et. al. [11] introduced Active Neural Localization which predicts a likelihood map for the agent location on the map and uses such information to predict its policy. The main advantage of using RL is that it doesn't need manual labeling. Imitation Learning based models represent another direction for navigation in literature where the models learn to mimic the human behavior. Ross et. al. [12] were able to produce on-board UAV model to avoid obstacles in the forest using the DAgger algorithm [13] which is one of the most used Imitation Learning algorithms. The trained model depends on differnt kind of visual features from the input images. Kelchtermans et. al. [14] used LSTM neural network and imitation learning to perform the navigation based on a sequence of input images. Some work also used direct Supervised Learning for control. Giusti et. al. [15] developed a DL based model to control a drone in order to fly over forest trails. The actions produced by the network were discrete (go right - go straight - go left). Smolyanskiy et. al. [16] used similar model for trail following and added entropy reward to stabilize the drone navigation. Bojarski et. al. [8] used a CNN to map the input image of a paved road to a steering angle of the car, the final model has successfully learnt to associate the visual input of the road with the suitable angle to keep the car in the lane. Kim et. al. [17] trained a model for specific object search in indoor environment using CNN. In order to increase the model generalization, they augmented the training data and started the training with a pretrained model.

For producing a more destination specific navigation system, there are some work for path following. Brandão et. al. [18] developed a method for line following on water banks and similar patterns using Gaussian low-pass filter followed by moving average smoothing on the received input image to control UAVs. De Mel et. al. [19] used optical flow between consecutive frames to calculate the relative position of the UAV instead of GPS. Nguyen et. al. [20] used the Funnel Lane theory and extend it to be used for controlling the drone's yaw and height using Kanade-Lucas-Tomasi (KLT) corner features.

On the other hand, having a separate localization system can help the navigation to be more accurate. Karim et. al. [21] designed a global localization system by classifying the district above which the drone is flying. Wang et. al. [22] developed an end to end system for calculating the odometry using visual information. Clark et. al. [23] proposed adding inertial information from IMU to visual information from onboard camera to localize a drone in an indoor environment. Kendall et. al. [24] trained a CNN to predict the camera location and orientation directly from a single input image in outdoor scene with landmarks. Melekhov et. al. [25] calculated the relative pose from two images by fine tuning a pretrained CNN.

### 3. Proposed Approach

#### 3.1. Methodology

Our approach uses a CNN to output the yaw angle by which the drone should rotate. We used a pretrained VGG-16 network [26]s, we removed the softmax layer and all fully connected layers as these layers are trained for the classification task in ILSVRC [2], then we inserted a fully connected layer containing 512 hidden units followed by another fully connected layer that outputs the yaw angle acting as a regressor function. The input is normalized to have zero mean and unit standard deviation because the VGG-16 network is trained with such normalization, the weights of the convolutional layers are frozen to increase the training speed, and also we did not want the convolutional layers to extract features that are biased to the synthetic simulated environments as eventually we want to proof our idea of visual path following based on a generic feature extractor and a regressor in a real world setting.

The network is trained end-to-end to give the yaw angle which controls the drone navigation based on the visual input of the current scene input only. We trained the regressor using Adam [27] optimizer with learning rate 1e-4 and batch size 64 for 100 epochs, the learning rate is divided by 2 every 25 epochs. The loss function is the Mean Squared Error (MSE) between the predicted yaw and the true yaw. The true yaw is calculated as the angle between the next waypoint and the drone's heading yaw that is the angle by which the drone is deviated from its next waypoint.

We used this loss to help the network correlate between the visual input, waypoints of the path, and the deviation angle between that visual input and the next waypoint. To train for the whole path we divided it into independent segments assuming that waypoints will not overlap and that each unique path will have a separate model that is trained to follow it.

#### 3.2. Control

For simplicity, we used a fixed height during our experiments. In training and testing the drone moves by a fixed step size of value 0.2. In training, the direction of the step size is determined by the optimal shortest direction between the drone's current position and the position of the next waypoint. The optimal shortest direction is the vector joining the drone's position and the next waypoint assuming that the predetermined path does not have obstacles between any two successive waypoints. In testing, the direction is determined by the CNN and the regressor. In the experiments we restricted the drone's movement to be in the direction of its heading.

#### 3.3. Path Augmentations

An important remark for the success of our approach is the use of path augmentations, as training for the optimal path only does not allow the model to explore other positions with different poses than those covered in the optimal path. Thus, if the model made a small drift it will result in a different visual scene from those that the model has been trained on. To mitigate this problem, we generated different augmented paths from the optimal path. Since we can calculate the optimal shortest direction to the next waypoint from the simulator, we add noise to both the position and the heading direction. The position is perturbed by adding a uniform random value between [-1, 1] meters to the optimal position, and the heading yaw is perturbed by adding a uniform random value between [-0.1, 0.1] radians.

Figure 3 shows the top-down view of the augmented paths in the Blocks environment, blue dots represents the original path, the red dot marks the starting position of the drone, the yellow and green lines mark the field of view (FOV) of the drone, the FOV of the drone is 90°. The following images in figure 3 show the augmented paths, noting that the full path is much denser than those images as shown in the last image in the same figure, those images are sparse just for the visualization to be clear.

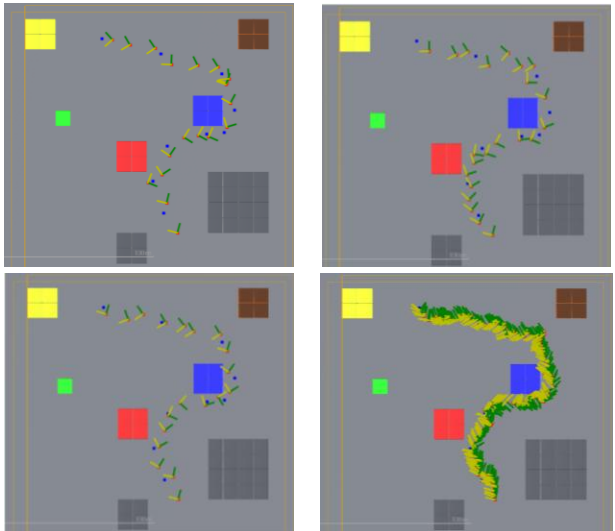

Figure 3. Three sparse augmented paths (first row and second row left), full dense augmented path (second row right).

### 3.4. Evaluation metrics

We have used two error metrics to compare the results of path following quantitatively. First, we loop on the desired path's waypoints and calculate the minimum distance that is approached by the drone to each waypoint and average over all the waypoints, we call this metric Mean Waypoints Minimum Distance. The second metric we used is the shortest distance between the drone's location and the next two closest waypoints, we call this metric Mean Cross Track Distance.

We also used a third metric to estimate the difficulty of the path in terms of change of angles, we loop on the desired path's waypoints and calculate the angle between each two successive waypoints, then accumulate the sum of the difference of these angles. We call this metric the Sum of Angle Change.

### 4. Dataset

To collect visual data we used Unreal Engine 4 integrated with AirSim plugin [1] that provided the API for the drone control and for data collection. We generated two paths in the Blocks environment and two paths in the Landscape Mountains environment. As mentioned in section 3, each path is augmented by adding noise to the optimal shortest direction to the next waypoint. For each unique path we generated 16 augmented paths from the optimal path and used images from those 16 augmented paths in training, noting that the model does have any information about the order of waypoints or the sequence of the input as each image is fed separately to the model and each image should be correlated with a yaw heading to keep the drone in the specified path.

Training for different paths is done separately, thus for each path there is a network that has learnt to follow that path by correlating the image with the yaw heading that is required for the next waypoint. We did not try joint training of different paths as each path is conditioned on its starting point, and different paths could possibly have conflicting decisions when faced with the same visual input. Table 1 shows the number of images per each path after augmentation with the 16 jittered paths, images are of the same size 512x288. Distance of the path is measured as the summation of the Euclidean distance between each two successive waypoints.

| Path ID | Total number of train images | Distance (meters) | Sum of Angle Change (Radians) |
|---|---|---|---|
| Path 1 (Blocks) | 6824 | 145.90 | 5.00 |
| Path 2 (Blocks) | 19490 | 239.39 | 4.48 |
| Path 3 (Landscape Mountains) | 20993 | 267.22 | 4.23 |
| Path 4 (Landscape Mountains) | 32364 | 412.61 | 6.63 |

Table 1. Number of images per path

## 5. Experimental Results

### 5.1. Path Augmentations

We firstly tested our approach after training the network on one path only without augmentation. Figure 4 shows the top-down view of the desired path to be followed in the left image, and the path covered by the drone in the right image. As shown in the figure, the drone's path has drifted from the desired path after sometime because the network did not output the correct yaw during the turn. Once the drone had a small drift it will not be able to recover the true path as it had not seen these parts of the scene during training. As mentioned in section 3.3, this problem can be alleviated by using augmented jittered paths that are generated from the true optimal path during training.

To furtherly test the effect of the number of augmented paths, we generated another four augmented test paths that are used to quantify the angle MSE predicted by the network for each image in the four test paths. Figure 4 bottom image shows the test error when using (1, 4, 8, 16) augmented paths in training, as the number of augmented paths increases the test error decreases reaching a minimum value of (0.0122, 0.0121, 0.0116, 0.01009) radians squared respectively. As expected increasing the number of augmented paths reduces the test angle MSE. Though the absolute difference value between them is small, in a real path following test, angle errors would accumulate which would result in losing track of the desired path. Figure 4 shows the deployed network after being trained on (1, 4, 8, 16) augmented paths, (1, 4) has drifted and failed to recover the desired path. 8 augmented paths has made a small drift then recovered. After training on 16 paths, the network has followed the desired path as shown in the figure with minimal drift.

approach in different scenarios. All of the next experiments have used 16 augmented paths. Figure 5 shows top-down view of path 2 to be followed in the Blocks environment, the network has succeeded to follow the path.

To test our approach on more realistic scenes, we generated two paths in the Landscape Mountains environment. Figure 6 shows top-down view of path 3 and path 4, the network has succeeded to follow both paths.

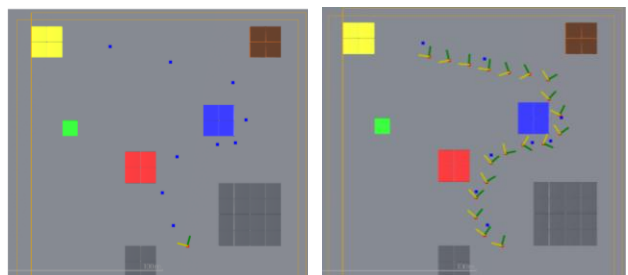

Figure 5. Original path 2 (left), followed path (right).

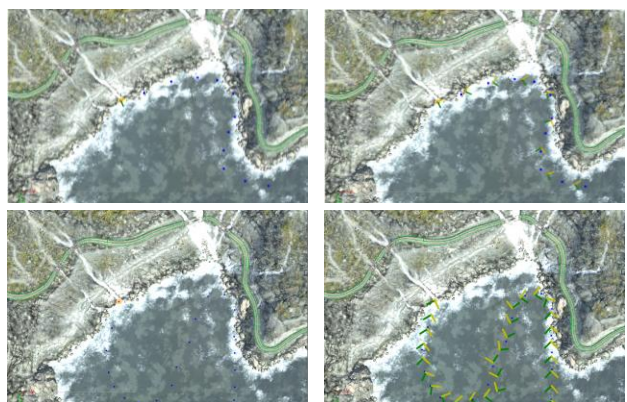

Figure 6. Path 3 and path 4, original (left), followed path (right).

### 5.3. Quantitative Evaluation

Table 2 shows the Mean Waypoints Minimum Distance and the Mean Cross Track Distance metrics mentioned in section 3.4 to each one of the four paths and also the average of these four paths.

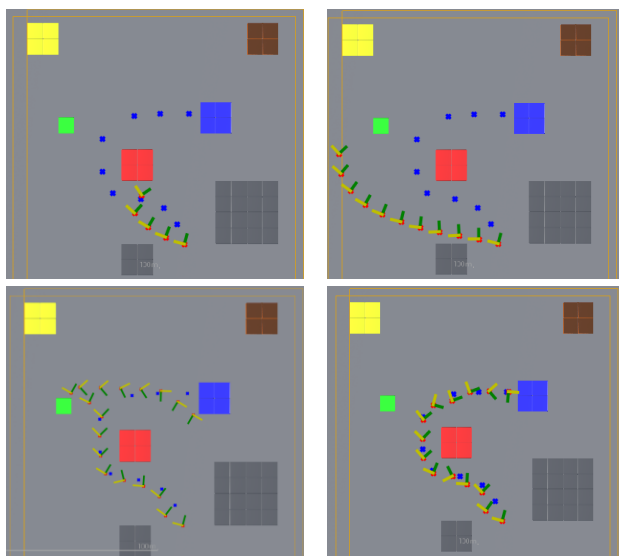

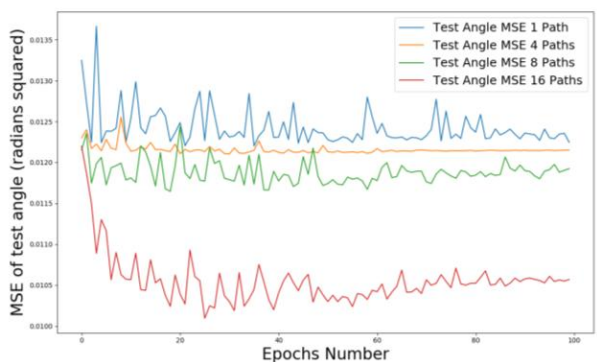

Figure 4. Followed paths after training on (1, 4, 8, 16) paths in first and second row for path 1, angle MSE loss in third row.

### 5.2. Different Paths

Path 1 in Blocks environment is shown in figure 4. We have generated different paths to be followed to test our

| Path ID | Mean Waypoints Minimum Distance (meters) | Mean Cross Track Distance (meters) |
|---|---|---|
| Path 1 (Blocks) | 5.20 | 5.41 |
| Path 2 (Blocks) | 4.85 | 3.70 |
| Path 3 (Landscape Mountains) | 0.56 | 0.69 |
| Path 4 (Landscape | 1.14 | 1.72 |

| Mountains) | | |
|---|---|---|
| Average across paths | 2.94 | 2.88 |

Table 2. Evaluation results of four paths

Results show that Landscape Mountains environment has smaller error compared to the Blocks environment, this is possibly because Landscape Mountains contain richer visual features that could be used for navigation and also the pretrained VGG-16 model was trained on images that contain natural images which is more similar to Landscape Mountains than to the Blocks environment.

The results are consistent with the Sum of Angles Change metric that is mentioned in section 3.4. Path 1 has larger distance error than path 2 possibly due to having larger Sum of Angles change, and path 4 also has larger distance error than path 3 due to having larger Sum of Angles change. This is attributed to paths having larger angle changes would typically have larger drifts and distance errors.

## 6. Conclusions

In this work we presented using visual input for path following in GPS denied environments. We have shown that a CNN can successfully predict the steering angle required to move the drone to the next waypoint in its path, and hence follow the whole path. An average of 2.88 meters cross track distance has been achieved across four paths. In future work, we intend to implement our work on a real drone to solidify the success of our approach in a real-world setting. We are also working to incorporate the time information to help the model with overlapping paths as well as using a sequence of input frames instead of using the current drone's image only. Future work can also include integrating the proposed end-to-end navigation system in UAV middleware [28] as a standalone component and/or within a general framework for target detection and tracking [29][30] that builds on our prior work in these areas.

## 7. References

[1] S. Shah, D. Dey, C. Lovett, and A. Kapoor, "AirSim: High-Fidelity Visual and Physical Simulation for Autonomous Vehicles," in *Field and Service Robotics*, 2017.

[2] Jia Deng, Wei Dong, R. Socher, Li-Jia Li, Kai Li, and Li Fei-Fei, "ImageNet: A large-scale hierarchical image database," *2009 IEEE Conf. Comput. Vis. Pattern Recognit.*, pp. 248–255, 2009.

[3] P. Sermanet, D. Eigen, X. Zhang, M. Mathieu, R. Fergus, and Y. LeCun, "OverFeat: Integrated Recognition, Localization and Detection using Convolutional Networks," *CoRR*, vol. abs/1312.6229, 2013.

[4] J. Redmon, S. Divvala, R. Girshick, and A. Farhadi, "（2016 YOLO） You Only Look Once: Unified, Real-Time Object Detection," *Cvpr 2016*, pp. 779–788, 2016.

[5] R. Girshick, "Fast R-CNN," in *Proceedings of the IEEE International Conference on Computer Vision*, 2015, vol. 2015 Inter, pp. 1440–1448.

[6] V. Badrinarayanan, A. Kendall, and R. Cipolla, "SegNet: A Deep Convolutional Encoder-Decoder Architecture for Image Segmentation," *IEEE Trans. Pattern Anal. Mach. Intell.*, vol. 39, no. 12, pp. 2481–2495, 2017.

[7] O. Ronneberger, P. Fischer, and T. Brox, "U-Net: Convolutional Networks for Biomedical Image Segmentation," *Miccai*, pp. 234–241, 2015.

[8] M. Bojarski *et al.*, "End to End Learning for Self-Driving Cars," pp. 1–9, 2016.

[9] T. Zhang, G. Kahn, S. Levine, and P. Abbeel, "Learning Deep Control Policies for Autonomous Aerial Vehicles with MPC-Guided Policy Search," 2015.

[10] Y. Zhu *et al.*, "Target-driven Visual Navigation in Indoor Scenes using Deep Reinforcement Learning," 2016.

[11] D. S. Chaplot, E. Parisotto, and R. Salakhutdinov, "Active Neural Localization," pp. 1–15, 2018.

[12] S. Ross *et al.*, "Learning monocular reactive UAV control in cluttered natural environments," *Proc. - IEEE Int. Conf. Robot. Autom.*, pp. 1765–1772, 2013.

[13] S. Ross, G. Gordon, and D. Bagnell, "A reduction of imitation learning and structured prediction to no-regret online learning," in *Proceedings of the fourteenth international conference on artificial intelligence and statistics*, 2011, pp. 627–635.

[14] K. Kelchtermans and T. Tuytelaars, "How hard is it to cross the room? -- Training (Recurrent) Neural Networks to steer a UAV," 2017.

[15] A. Giusti *et al.*, "A Machine Learning Approach to Visual Perception of Forest Trails for Mobile Robots," *IEEE Robot. Autom. Lett.*, vol. 1, no. 2, pp. 661–667, 2016.

[16] N. Smolyanskiy, A. Kamenev, J. Smith, and S. Birchfield, "Toward low-flying autonomous MAV trail navigation using deep neural networks for environmental awareness," *IEEE Int. Conf. Intell. Robot. Syst.*, vol. 2017–Septe, pp. 4241–4247, 2017.

[17] D. K. Kim and T. Chen, "Deep Neural Network for Real-Time Autonomous Indoor Navigation," 2015.

[18] A. Brandão, F. Martins, and H. Soneguetti, "A Vision-based Line Following Strategy for an Autonomous UAV," *Proc. 12th Int. Conf. Informatics Control. Autom. Robot.*, pp. 314–319, 2015.

[19] D. H. S. De Mel, K. A. Stol, J. A. D. Mills, and B. R. Eastwood, "Vision-based object path following on a quadcopter for GPS-denied environments," *2017 Int. Conf. Unmanned Aircr. Syst. ICUAS 2017*, pp. 456–461, 2017.

[20] T. Nguyen, G. K. I. Mann, and R. G. Gosine, "Vision-based qualitative path-following control of quadrotor aerial vehicle," *2014 Int. Conf. Unmanned Aircr. Syst. ICUAS 2014 - Conf. Proc.*, pp. 412–417, 2014.

[21] K. Amer, M. Samy, R. ElHakim, M. Shaker, and M. ElHelw, "Convolutional Neural Network-Based Deep Urban Signatures With Application to Drone Localization," in *The IEEE International Conference on Computer Vision (ICCV)*, 2017.

[22] H. Wen, S. Wang, R. Clark, and N. Trigoni, "{DeepVO}: Towards End to End Visual Odometry with Deep Recurrent Convolutional Neural Networks," *Proc. {IEEE} Int. Conf. Robot. Autom.*, pp. 2043–2050, 2017.

[23] R. Clark, S. Wang, H. Wen, A. Markham, and A. Trigoni, "VINet: Visual-Inertial Odometry as a Sequence-to-Sequence Learning Problem," in *AAAI*, 2017.

[24] A. Kendall, M. Grimes, and R. Cipolla, "Posenet: A convolutional network for real-time 6-dof camera relocalization," in *Computer Vision (ICCV), 2015 IEEE International Conference on*, 2015, pp. 2938–2946.

[25] I. Melekhov, J. Ylioinas, J. Kannala, and E. Rahtu, "Relative Camera Pose Estimation Using Convolutional Neural Networks," in *Advanced Concepts for Intelligent Vision Systems*, 2017, pp. 675–687.

[26] K. Simonyan and A. Zisserman, "Very deep convolutional networks for large-scale image recognition," *arXiv Prepr. arXiv1409.1556*, 2014.

[27] D. P. Kingma and J. Ba, "Adam: A method for stochastic optimization," *arXiv Prepr. arXiv1412.6980*, pp. 1–15, 2014.

[28] A. El-Sayed and M. ElHelw, "Distributed component-based framework for unmanned air vehicle systems", Proceedings of the IEEE International Conference on Information and Automation, 2012.

[29] A. Salaheldin, S. Maher, M. Helw. "Robust real-time tracking with diverse ensembles and random projections", Proceedings of the IEEE International Conference on Computer Vision (ICCV) Workshops, 2013, pp. 112-120.

[30] M. Siam, M. Elhelw. "Enhanced target tracking in UAV imagery with PN learning and structural constraints", Proceedings of the IEEE International Conference on Computer Vision (ICCV) Workshops, 2013, pp. 586-593.